\documentclass[10pt,twocolumn,letterpaper]{article}
\usepackage{url}
\usepackage{cvpr}
\usepackage{times}
\usepackage{epsfig}
\usepackage{graphicx}
\usepackage{amsmath}
\usepackage{amssymb}
\usepackage{capt-of}
\usepackage{verbatim}
\usepackage{threeparttable}
\usepackage[linesnumbered,lined,ruled]{algorithm2e}

\usepackage[breaklinks=true,bookmarks=false]{hyperref}

\cvprfinalcopy 

\def\cvprPaperID{2494} 

\begin{document}

\title{RENAS: Reinforced Evolutionary Neural Architecture Search}

\author{\\ \textbf{Yukang Chen$^{1,2}$\quad Gaofeng Meng$^{1,2}$\quad Qian Zhang${}^{3}$\quad Shiming Xiang$^{1,2}$} \\
 \textbf{\quad Chang Huang${}^{3}$ \quad Lisen Mu${}^{3}$
\quad Xinggang Wang${}^{4}$}\\
${}^{1}$National Laboratory of Pattern Recognition, Institute of Automation, Chinese Academy of Sciences\\ 
${}^{2}$School of Artificial Intelligence, University of Chinese Academy of Sciences\\ 
${}^{3}$Horizon Robotics\quad ${}^{4}$ Huazhong University of Science and Technology  \\
\emph{\{yukang.chen,gfmeng,smxiang\}@nlpr.ia.ac.cn}  \\ 
\emph{\{qian01.zhang,chang.huang,lisen.mu\}@horizon.ai, \{xgwang\}@hust.edu.cn}
}
\maketitle

\begin{abstract}
   Neural Architecture Search (NAS) is an important yet challenging task in network design due to its high computational consumption. 
  To address this issue, we propose the Reinforced Evolutionary Neural Architecture Search (RENAS), which is an evolutionary method with reinforced mutation for NAS. 
	Our method integrates reinforced mutation into an evolution algorithm for neural architecture exploration, in which a mutation controller is introduced to learn the effects of slight modifications and make mutation actions. The reinforced mutation controller guides the model population to evolve efficiently.
	Furthermore, as child models can inherit parameters from their parents during evolution, our method requires very limited computational resources.
	In experiments, we conduct the proposed search method on CIFAR-10 and obtain a powerful network architecture, RENASNet. This architecture achieves a competitive result on CIFAR-10. The explored network architecture is transferable to ImageNet and achieves a new state-of-the-art accuracy, i.e., 75.7\% top-1 accuracy with 5.36M parameters on mobile ImageNet.
		We further test its performance on semantic segmentation with DeepLabv3 on the PASCAL VOC. RENASNet outperforms MobileNet-v1, MobileNet-v2 and NASNet. It achieves 75.83\% mIOU without being pretrained on COCO. The code for training and evaluating ImageNet models is available at {\small \color{red} \url{https://github.com/yukang2017/RENAS}}.
\end{abstract}
\begin{figure*}
  \centering
  \includegraphics[width=1.0\textwidth]{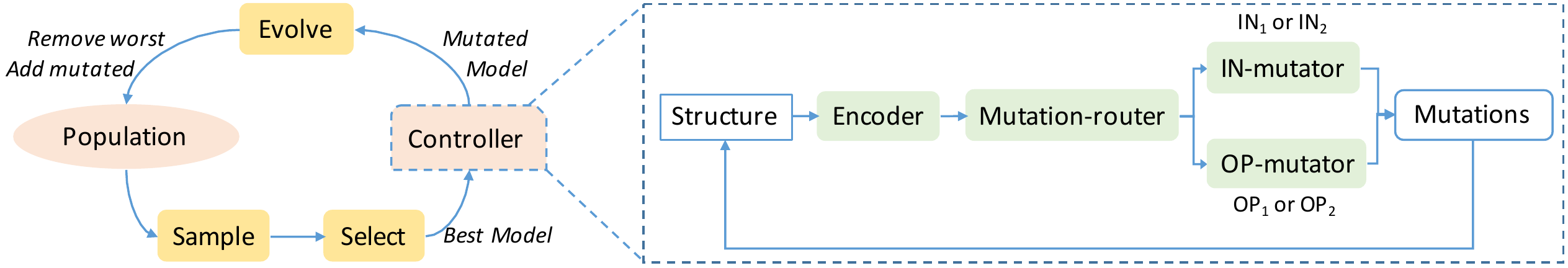}
  \caption{The evolutionary neural architecture search framework and the structure of the reinfored mutation controller.}\label{fig:1}
\end{figure*}

\section{Introduction}
Recent several years have witnessed the great success of neural networks~\cite{szegedy2015going,he2016deep,szegedy2016rethinking,szegedy2017inception,huang2017densely} in tackling various challenging tasks, e.g., image classification, object detection and semantic segmentation.
 However, designing hand-crafted neural networks is still a laborious task due to the heavy reliance on expert experience and large amount of trials. For example,
 hundreds of experts in academia and industry have made great efforts to optimize the architectures of neural networks that increase the top-5 accuracy to 96.43\% on the ImageNet challenge from AlexNet~\cite{krizhevsky2012imagenet}, VGG~\cite{simonyan2014very}, Inception~\cite{szegedy2015going} to ResNet~\cite{he2016deep}. 
 
Techniques in automated network architecture design have attracted increasing research interests.
Many neural architecture search methods 
 have been proposed and proven to be capable of yielding high-performance models.
A large portion of these methods are based on Reinforcement Learning (RL)~\cite{zoph2017learning,cai2018efficient,zoph2016neural}. Typical RL-based NAS methods construct networks sequentially, e.g., by using a RNN controller~\cite{zoph2017learning,pham2018efficient,zoph2016neural} to determine a sequence of operator and connection tokens.
In addition to RL, Evolution Algorithm (EA) is also employed in many works~\cite{Real2018Regularized,stanley2002evolving,real2017large,miikkulainen2017evolving,xie2017genetic}.
In EA-based NAS methods, a population of architectures are initialized first and then evolved with their validation accuracies as fitnesses. 



EA and RL have achieved the state-of-the-art performance in the task of NAS. 
However, both of them have limitations respectively:
1) For EA-based NAS, it tends to evolve a population of architectures that guarantees the diversity of potential results. However, as the evolution progress relies heavily on random uncontrollable mutation, the efficiency of EA has no guarantee. For instance, AmoebaNet~\cite{Real2018Regularized} is searched by an EA-based method and has better final results than its RL counterpart, NASNet~\cite{zoph2017learning}. But in the same search space, AmoebaNet~\cite{Real2018Regularized} uses more computational resources than NASNet~\cite{zoph2017learning} (3150 GPU days vs 2000 GPU days).
2) For RL-based NAS, it relies on hyper-parameters to guarantee stability.
But when determining an architecture layer by layer, RL controller needs to try tens of actions to get a positive reward as a supervisory signal. This makes the training process inefficient.

In this paper, we propose the Reinforced Evolutionary Neural Architecture Search (RENAS), which integrates RL into the evolution framework to address the above issues. Our method introduces a reinforced mutation controller to help the efficient exploration of the search space. Thanks to the nature of EA, the child model could inherit most parameters from its parent, which in turn makes the search more efficient. Our main contributions are summarized as below:



\begin{enumerate}
\item[-] A novel neural architecture search framework is proposed with EA and RL integrated. This framework integrates the advantages of both of them and ensures the search efficiency.

\item[-] We design a reinforced mutation controller to learn the effects of slight modifications and make actions to guide the evolution. This technique helps the population evolve to a better status in fewer iterations.
\item[-] A powerful neural architecture, RENASNet, is discovered. It achieves a competitive accuracy on CIFAR-10, i.e., 2.88\% $\pm$ 0.02 and a new state-of-the-art on mobile ImageNet with 75.7\% top-1 accuracy and 5.36M parameters. We further test its performance on semantic segmentation with DeepLabv3~\cite{deeplab-v3} on the PASCAL VOC 2012~\cite{PascalVOC}. RENASNet outperforms the state-of-the-art networks and achieves 75.83\% mIOU without being pretrained on COCO~\cite{COCO}.
\end{enumerate}

\section{Related Work}
\subsection{RL-based NAS}
Reinforcement learning gains much research attention in recent works~\cite{zoph2017learning,cai2018efficient,zoph2016neural,baker2016designing}. In NAS~\cite{zoph2016neural}, neural networks are specified by variable-length strings which are generated by a RNN controller. The network specified by a string is then trained to return a validation accuracy. In turn, the controller is updated with policy gradient using the accuracy as reward. In this framework, networks specified by strings are generated layer by layer. The success reported by NAS~\cite{zoph2016neural} inspires many other valuable works, but the expensive computational cost (28 days with 800 GPUs) limits its wide application, since training and evaluating a single model is time-consuming. 

\subsection{EA-based NAS}
Evolution process in the nature is intuitively similar to NAS. Thus, many early automatic architecture search methods~\cite{miller1989designing,DBLP:journals/tnn/YaoL97,stanley2002evolving,real2017large,miikkulainen2017evolving,xie2017genetic} adopt EA to evolve a population of models. For instance, the large scale evolutionary method~\cite{real2017large} explores a CNN search space with neuro-evolution algorithm, which returns networks matching the human-designed models. The framework of our paper is based on AmoebaNets~\cite{Real2018Regularized}, in which a common evolutionary algorithm, tournament selection strategy, matches or even outperforms its RL baseline~\cite{zoph2017learning} in speed and accuracy. However, evolution process is slow due to the random mutation. To address this issue, we introduce a controller for mutation to guide the evolution process.

\subsection{Efficient NAS}
Difficulties of NAS mainly come from the extremely large search space and the time-consuming model evaluation. In NASNet~\cite{zoph2017learning}, computational cost is saved with cell-wise search space, which is adopted by the following works~\cite{Real2018Regularized,pham2018efficient,liu2017progressive}. Instead of exploring the whole network architecture, NASNet~\cite{zoph2017learning} centers on learning cell structures which are then stacked multiple times into a complete network, making the output networks scalable for various datasets and tasks.
In addition, a variety of techniques on accelerating evaluation have proven effective: Block-QNN~\cite{zhong2018practical} improves the search speed with an early-stop strategy. ENAS~\cite{pham2018efficient} utilizes parameter sharing among child models instead of training from scratch. EAS~\cite{cai2018efficient} utilizes the Net2Net transformation~\cite{chen2015net2net} to reuse parameters. Accuracy prediction, used in this work~\cite{baker2018accelerating}, is also a novel technique to save computational resources, although the accuracy predictor might not always be accurate enough. 

 \begin{figure*}[ht]
  \centering
  \includegraphics[width=1.0\textwidth]{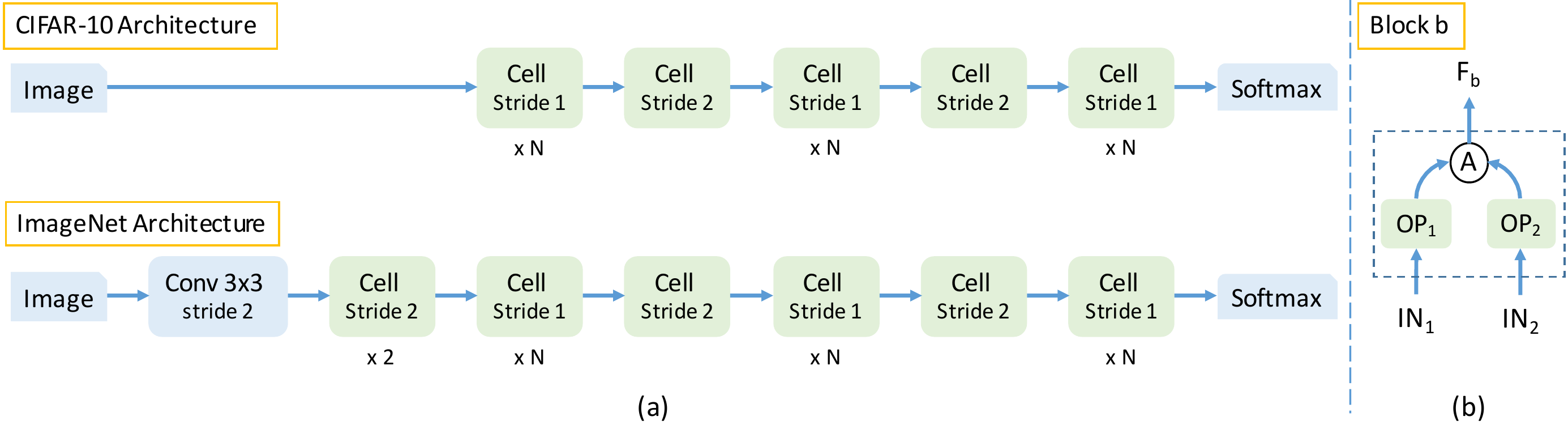}
  \caption{(a) Architectures employed for CIFAR-10 and ImageNet datasets respectively. During searching, the network structure is specified by the cell structure. 
The image size in ImageNet (224x224) is much larger than that in CIFAR-10 (32x32). So there are additional reduction cells and convolution 3x3 with stride 2 in ImageNet architectures to downsample feature maps.  
   $\;$
  (b) Each cell consists of \texttt{\#B} blocks. Each block takes in two inputs $\{i_1,i_2\}$, apply specific operations $\{o_1,o_2\}$ to them respectively and then combine them with element-wise addition to generate a feature map $O_b$. We search for $\{i_1,i_2,o_1,o_2\}$ for \texttt{\#B} blocks to construct a reasonable cell structure, which in turn constitutes a network.}\label{fig:2}
\end{figure*}

\subsection{Integration of EA and RL}
RL has shown its capacity of accelerating evolutionary progress via Baldwinian or Lamarckian mechanisms~\cite{downing2001adaptive}.
The idea of Integration RL and EA has been previously investigated, but our method is distinctly different from previous works.
In~\cite{DBLP:journals/gpem/Downing01}, RL is used to enhance standard tree-based genetic programming in maze problems. In~\cite{kamio2005adaptation}, integration of EA and RL is used to improve the adaption actions of a real robot. In~\cite{hitoshi1998multi}, EA is integrated into a multi-agent Q-learning to shrink the search space. 

In our method, a mutation controller is integrated into the evolution framework to learn the effects of modifications and make reasonable mutation actions. Compared to only RL methods and only EA methods, this integration brings us the following benefits:

1) RL training becomes more efficient. Because making modifications to a network needs much fewer actions to make than constructing a model layer by layer.  As the child model is modified from the parent model, the mutation controller is easy to learn the effects of slight differences.

2) The evolution process becomes more efficient and stable with the help of the reinforced mutation controller. Model architectures and their fitnesses (validation accuracies) are previously neglected but valuable hints generated during evolution. We reuse these useful supervisory signals to train the mutation controller. It in turn eliminates the accumulation of harmful mutation.

\section{Search Space}
\label{3}
Rather than designing the entire convolutional network, we adopt the idea that learning cell structures~\cite{zoph2017learning}.
In this section, we introduce the search space by factorizing each network  into cells and blocks. 
The architecture frames and the inner block are illustrated in Fig.~\ref{fig:2}.

\subsection{Block} 
Each block maps two inputs into one output feature map as shown in Fig.~\ref{fig:2} (b). It takes two input feature maps $\{i_{1}, i_{2}\}$, applies two operators $\{o_{1}, o_{2}\}$ to them respectively and then combines them into an output $O$ via element-wise addition $A$.
For this reason, each block could be specified by a string of length 4, $\{i_1, i_2, o_1, o_2\}$. 
$\{i_1,i_2\}$ are selected from $\{O_1^c,O_2^c, ... ,O_{b-1}^c, O_B^{c-1}, O_B^{c-2}\}$, where $O_1^c, ... ,O_{b-1}^c$ are outputs of previous blocks in the current cell. $O_B^{c-1}$ and $O_B^{c-2}$ are outputs of the first and second previous cells. Operation choices for $\{o_1, o_2\}$ are selected from a set of 6 functions: 
3x3 depth-wise separable convolution, 
5x5 depthwise-separable convolution, 
7x7 depth-wise separable convolution, 
3x3 avg pooling, 3x3 max pooling, identity.

\subsection{Cell}
Each cell can be represented as a directed acyclic graph which consists of \texttt{\#B} blocks. Assume there is an input feature map with shape $h\times w \times f$, where $h$ and $w$ denote the height and width of the feature and $f$ means the channel number. Cells with stride 2 output features with shape $\frac{h}{2} \times \frac{w}{2} \times 2f$ while cells with stride 1 keep the shape of feature maps. Each cell consist of \texttt{\#B} blocks. Therefore, we search for the structure of each block and how they connect together to build a cell.

\subsection{Network}
Each network could be specified with three factors: the cell structure, \texttt{\#N} the number of cells to be stacked and \texttt{\#F} the number of filters in the first layer. As we fix \texttt{\#N} and \texttt{\#F} during search, our search space is constrained to all possible cell structures. Once search finished, models are constructed with different sizes to fit various tasks or datasets. We adjust the number of cells repeated \texttt{\#N} and the number of filters in the first layer to control the depth and width of networks. As illustrated in Fig.~\ref{fig:2} (a), the architecture for ImageNet has two more cells with stride 2 and a deeper steam. Because the image size in ImageNet (224x224) is much larger than that in CIFAR-10 (32x32), it needs more downsample operations.

Each network therefore is specified with \texttt{5\#B} tokens, \texttt{4\#B} of which are variable during search. Because each cell consists of \texttt{\#B} blocks and each block is specified by 5 tokens: two inputs $\{i_{1}, i_{2}\}$, two operations $\{o_{1}, o_{2}\}$ and a combination operation $A$ that is fixed as addition.
 Therefore, searching network architecture is converted to search for \texttt{4\#B} variables.
This search space is smaller than NASNet search space~\cite{zoph2017learning}.
We use only one cell type and reduce the feature map size using cells with stride 2. Besides we use 6 candidate functions and fix combiners as element-wise addition.
The complexity of the search space can be estimated with ease. Each block consists of 2 nodes. For each node we need to select its input from $b + 1$ possible indexes and its operator from these 6 functions. As we set \texttt{\#B=5}, there are $(6^5 \times (5+1)!)^2 = 3.1 \times 10^{13}$ possible networks, which is still an extremely large  space.
\begin{algorithm}[t]
\caption{The framework of RENAS} \label{algo:1}
\SetKwInOut{Input}{input}\SetKwInOut{Output}{output}
\Input{num blocks each cell \texttt{\#B}, max num epochs \texttt{\#E}, num filters in first layer \texttt{\#F}, num cells in model \texttt{\#N}, population size \texttt{\#P}, sample size \texttt{\#S}, training set \texttt{$D_{train}$}, validation set \texttt{$D_{val}$}}

\Output{a population of models $P$}
\begin{small}
\texttt{$P^{(0)}$ $\gets$ initialize(\#F, \#N, \#P)}


\For{i=1:\#E}{

		\texttt{$S^{(i)}$ $\gets$ sample($P^{(i-1)}$, \#S)}
		
		\texttt{$B$,$W$ $\gets$ select($S^{(i)}$)}
		
		\texttt{$C$,  $\omega^{B}$ $\gets$ reinforced-mutate($B$)}
		
		 \texttt{$\omega^{C}$ $\gets$ finetune($C$,  $\omega^{B}$, $D_{train}$)}
		
		\texttt{$f_{C}$ $\gets$ evaluate($C$, $\omega^{C}$, $D_{val}$)}
				
		\texttt{$P^{(i)}$ $\gets$ push-pop($P^{(i-1)}$, $C$, $f_{C}$, $W$)}
}
\end{small}
\end{algorithm}

\section{Search Strategy}
\label{4}
\subsection{Evolution Framework}
\label{4.1}

 To search for architectures with high performance automatically, a population of models $\mathbf{P}$ is initialized randomly.
 Each \emph{individual} of $\mathbf{P}$ is trained on the training set $D_{train}$ and evaluated on the validation set $D_{val}$.
 Its fitness \emph{f} is defined as the validation accuracy. At each evolutionary step, a subset $\mathbf{S}$ is randomly sampled from $\mathbf{P}$.
 According to their fitnesses, the best individual \emph{B} and the worst individual \emph{W} are selected among $\mathbf{S}$.
 \emph{W} is excluded from $\mathbf{P}$ and \emph{B} becomes a \emph{parent} to produce a child \emph{C} with \emph{mutation}.
  \emph{C} is then trained and evaluated to measure its fitness $f_C$. Afterwards \emph{C} is put into $\mathbf{P}$. This scheme actually belongs to \emph{tournament selection}~\cite{goldberg1991comparative}, repeating competitions in random samples. The procedure is formulated in Algorithm~\ref{algo:1}. 

\subsection{Reinforced Mutation}
\label{4.2}
The reinforced mutation is implemented with a mutation controller to learn the effects of slight modifications and make mutation actions.
Fig.~\ref{fig:1} shows the framework of our controller, which implements a mechanism of attention. The controller takes a string of \texttt{5\#B} length which represents the given cell architecture. Specifically, our controller consists of 4 parts:
(1) an Encoder (\texttt{Enc}) following an embedding layer to learn the effect of each part of the cell,
(2) a Mutation-router (\texttt{Mut-rt}) to choose one from $i_1,i_2,o_1,o_2$ of the block,
(3) an Input-mutator (\texttt{IN-mut}) to change node's input with a new input $i_{new}$
(4) an OP-mutator (\texttt{OP-mut}) to change node's operator with a new operator $o_{new}$ .

\noindent
\textbf{Encoder}\quad
\texttt{Enc} is a bidirectional recurrent network with an input embedding layer. 
Hidden states learned by \texttt{Enc} indicate the effect of a local part on the whole network.
For block $b$ in \texttt{Enc}, its hidden states are $\{H_{i_1}^b, H_{i_2}^b, H_{o_1}^b, H_{o_2}^b\}$ where $H_{o_1}^b$ represents the effect of block $b$'s $o_1$ on the whole network. As each model is specified by \texttt{5\#B} numbers, \texttt{Enc} generates \texttt{5\#B} hidden states each step. Besides, we initialize two begin states, $H^{c-1},H^{c-2}$, which represent the information of the first and second previous cells.

For block $b$, the controller makes two decisions sequentially. At first, depending on $H_{i_1}^b, H_{i_2}^b, H_{o_1}^b, H_{o_2}^b$, \texttt{Mut-rt} decides which one of $i_1, i_2, o_1, o_2$ in block $b$ needs to be modified. It is sampled with a mechanism of attention via softmax classifiers. If one of input indexes, $i_1$ or $i_2$, is chosen, the \texttt{IN-mut} would be activated to pick one from  $\{O_1^c, ... ,O_{b-1}^c,O_B^{c-1},O_B^{c-2}\}$. Otherwise \texttt{OP-mut} would choose a new operator from that 6 operation choices. As there are B blocks in each cell, this process would be repeated for B times to modify a given architecture. Thus it makes \texttt{2\#B} modification actions for each model. We describe the implementation details in the following.

\noindent
\textbf{Mutation-router}\quad
\texttt{Mut-rt} is designed to find which ingredient of each block needs modification. For each block, \texttt{Mut-rt}'s inputs are a subset of \texttt{Enc}'s outputs $H_{i_1}^b, H_{i_2}^b, H_{o_1}^b, H_{o_2}^b$ and its output is one of  $i_1, i_2, o_1, o_2$, an $ID$ to mutate. We apply a fully connected layer to each hidden state use $softmax$ to compute the modification probability of each ingredient $P_{i_1}^b, P_{i_2}^b, P_{o_1}^b, P_{o_2}^b$ and sample one from $i_1, i_2, o_1, o_2$ with these probabilities.

\noindent
\textbf{IN-mutator}\quad
\texttt{IN-mut} chooses a new input for the node, if $ID\in (i_1,i_2)$. Its inputs include the chosen $ID$'s hidden state $H_{ID}^b$, the hidden states of all previous block's outputs $[H_{A}^1,...,H_{A}^{b-1}]$, and the hidden states of previous and previous-previous cells $H^{c-1},H^{c-2}$. We concat $[H_{A}^1,...,H_{A}^{b-1},H^{c-1},H^{c-2}]$ with $H_{ID}^b$ and apply a fully connected layer to them. Similar to \texttt{Mut-rt}, we use $softmax$ to compute the probability of replacing the original input with each substitute and then we determine $i_{new}$ by choosing from $1, ... ,b$-$1, c$-$1, c$-$2$ with these probabilities.

\noindent
\textbf{OP-mutator}\quad
\texttt{OP-mut} outputs an new operator $o_{new}$ depending on the input $H_{ID}^b$. This process is simple and similar to \texttt{Mut-rt}.

\begin{algorithm}[t]
\caption{Mutation generated by Controller} \label{algo:2}
\SetKwInOut{Input}{input}\SetKwInOut{Output}{output}
\Input{num blocks each cell \texttt{\#B}, a sequence $a$ of \texttt{4\#B} number specifying a cell}

\Output{a sequence of mutation actions $m$}
\begin{small}

$H^{c-1},H^{c-2}$ $\gets$ \texttt{Enc.begin()}

$H^1,...,H^B$  $\gets$ \texttt{Enc}$(H^{c-1},a)$

\For{b=1:\texttt{\#B}}{
		$H_{i_1}^b, H_{i_2}^b, H_{o_1}^b, H_{o_2}^b, H_{A}^b \gets H^b$
		
		$ID \gets$ \texttt{Mut-rt}$([H_{i_1}^b, H_{i_2}^b, H_{o_1}^b, H_{o_2}^b])$

		\eIf{$ID \in (i_1,i_2)$}{
				$i_{new} \gets$ \texttt{IN-mut} $(H_{ID}^b, [H_{A}^1,...,H_{A}^{b-1},H^{c-1},H^{c-2}])$
				
				$m^{(b)} \gets (ID,i_{new})$
		}{
				$o_{new} \gets$ \texttt{OP-mut} $(H_{ID}^b)$
				
				$m^{(b)} \gets (ID,o_{new})$
		}
}
\end{small}
\end{algorithm}

\subsection{Search Details}
\label{4.3}

\noindent
\textbf{Controller}\quad
At each evolution step, the controller makes a sequence of mutation actions. Then a child model $C$ is produced with the parent model modified. Then the validation accuracy $f_C$ is computed with parameter inheriting which is introduced in the following paragraph.
The reward $\gamma$ is a nonlinear function~\cite{cai2018efficient} of $f_C$, i.e., $\gamma=tan(f_C \cdot \frac{\pi}{2})$, since the gain of improving accuracy should be larger while the validation accuracy of its parent is higher.
The controller parameters $\theta$ is updated via policy gradient.

\noindent
\textbf{Child Models}\quad
Child models are trained and evaluated with its parameters inherited from their parents.
For each alive model $B$ in the population, we store its architecture string, its fitness $f_B$ and its learnable parameters $\omega^B$.
As each child model $C$ is generated from its parent model with slight modifications, differences between them only exist in the mutated layers. Therefore the child could inherit most parameters from the parent $B$. $\omega^C$ are classified into inheritable parameters $\omega^C_{inh}$ and new initialized parameters $\omega^C_{new}$.
And its fitness (validation accuracy) $f_C$ could be evaluated with fine-tuning instead of training from scratch. During fine-tuning, we train $\omega^C$ on a whole pass through $D_{train}$ with the learning rate of $\omega^C_{new}$ 10 times large as that of $\omega^C_{inh}$. In the experiments, the learning rate of $\omega^C_{new}$ equals to 0.01.

\noindent
\textbf{Deriving Architectures}\quad
During search, we set each cell contains \texttt{\#B=5} blocks, and \texttt{\#F=24} filters in the first convolution cell and we unroll the cells for \texttt{\#N=2}.
After the maximum number of epochs \texttt{\#E} is reached, we only retrain the models in the population from scratch and then take the model with highest accuracy. It is possible to improve our results by retraining more sampled models from scratch as done by other works~\cite{zoph2017learning,zoph2016neural}, but it is unfair to prove the performance of our controller. In the experiments, the population size \texttt{\#P} is set as 20.
For better comparison, we set \texttt{\#F} and \texttt{\#N} same to NASNets~\cite{zoph2017learning}.

\section{Experimental Results}
\label{5}
\begin{figure}[t]
  \centering
  \includegraphics[width=0.5\textwidth]{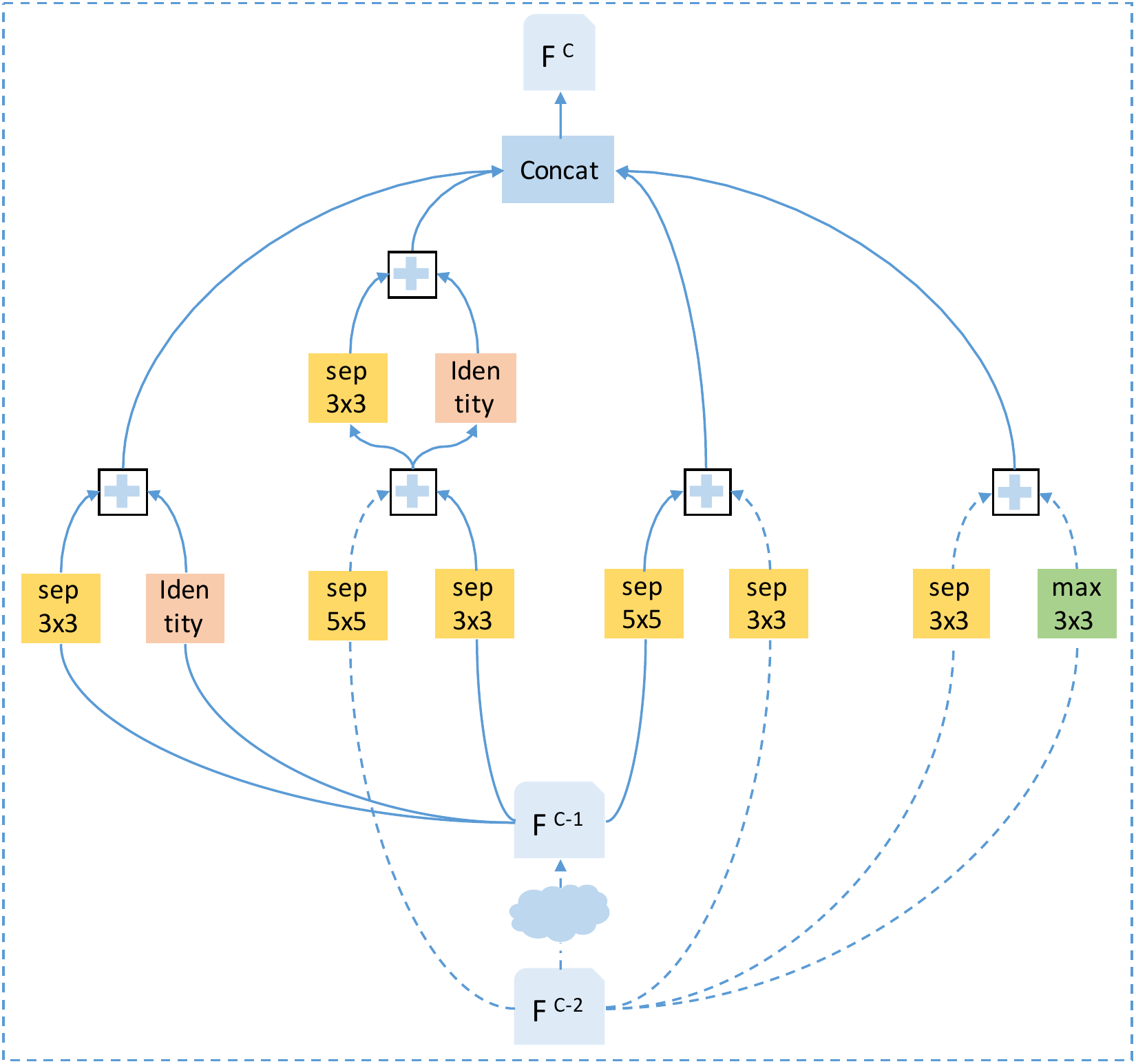}
  \captionof{figure}{The searched cell structure by RENAS. The full outer architectures for CIFAR and ImageNet are shown in Fig.~\ref{fig:2} (a).}
  \label{fig:3}
 \end{figure}

In this section, we first show our implementation details.
Then, we compare our searched architecture RENASNet (as illustrated in Fig.~\ref{fig:3}) with both state-of-the-art hand-design networks and other searched models on CIFAR-10 and ImageNet datasets. 
Ablation studies are made to show the search efficiency of RENAS. Further experiments show that RENASNet can be successfully transferred to achieve the semantic segmentation task.

  \begin{table*}[!hbt] 
  \centering  
    \caption{CIFAR-10 results. The top section presents the top hand-design networks, the middle section presents other architecture search results and the bottom section shows our results. \textbf{\#Params} means the number of free parameters.} \label{tab:1}
   \begin{threeparttable}
  \begin{tabular}{ l  c c c  c  c c}
  \hline
      \textbf{Model} 					     					& \textbf{Cutout}		& \textbf{GPUs} & \textbf{Days} & \textbf{\#Params} 	& \textbf{Error(\%)}  & \textbf{Method}\\  
      \hline
      DenseNet-BC~\cite{huang2017densely} 	& -			& - 						 &  - 		& 25.6M 							& 3.46 							  &  - \\  
      \hline
      PNASNet-5~\cite{liu2017progressive}		& -	 		& 100&1.5 						& 3.2M								& 3.41 $\pm$ 0.09 		 &		SMBO\\
      NASNet-A + cutout~\cite{zoph2017learning} 	& \checkmark & 500					&4 		& 3.3M								& 2.65 							 &		RL\\
	  AmoebaNet-B + cutout~\cite{Real2018Regularized} & \checkmark & 450&7 					& 2.8M 							& 2.55 $\pm$ 0.05 		 &		EA\\
	  ENAS + cutout~\cite{pham2018efficient}			& \checkmark & 1&0.5							& 4.6M								& 2.89 							 &		RL\\
	  DARTS (first order) + cutout~\cite{DBLP:journals/corr/abs-1806-09055} & \checkmark & 1&1.5		& 2.9M								&  2.94 & Gradient \\
	  DARTS (second order) + cutout~\cite{DBLP:journals/corr/abs-1806-09055} & \checkmark & 1&4		& 3.4M								&  2.83 $\pm$ 0.06 & Gradient\\
      \hline
      RENASNet (6, 32) + cutout									& \checkmark & 4&1.5							&	3.5M								&	2.88	$\pm$ 0.02		 &		EA\&RL\\
      \hline
  \end{tabular}
  
    \end{threeparttable}
   \end{table*}
  \begin{table*}[!hbt]
  \centering
  \caption{ImageNet classification results in the mobile setting. The results of hand-design models are in the top section, other NAS results are presented in the middle section and the result of our model is in the bottom section.} \label{tab:2}
      \begin{threeparttable}
  \begin{tabular}{ l  c  c  c c c}
  \hline
      \textbf{Model} 			      		 & \textbf{\#Params} & \textbf{\#Mult-Adds} & \textbf{Top-1/Top-5 Acc(\%)}  & \textbf{ Method}\\ 
      \hline  
	MobileNet-v1~\cite{howard2017mobilenets}			 & 4.2M		   &    569M			& 70.6 / 89.5	     & - \\
	MobileNet-v2 (1.4)\cite{sandler2018inverted}		 & 6.9M		   &	 585M			& 74.7 / -$\;\;\;\;\;$	   		  & -\\
	ShuffleNet-v1 2x~\cite{zhang2017shufflenet}			 &  $ \approx $ 5M			   &    524M			& 73.7 / -$\;\;\;\;\;$	    		& -\\
	ShuffleNet-v2 2x  (with SE)~\cite{DBLP:journals/corr/abs-1807-11164}			 & $ \approx $ 5M			   &    597M			& 75.4 / -$\;\;\;\;\;$	   & -\\

      \hline
	NASNet-A~\cite{zoph2017learning}		 & 5.3M		   &	 564M			& 74.0 / 91.6	  & RL\\
	NASNet-B~\cite{zoph2017learning}		 & 5.3M		   &	 488M			& 72.8 / 91.3	  & RL\\
	NASNet-C~\cite{zoph2017learning}		 & 4.9M		   &	 558M			& 72.5 / 91.0	  & RL\\
	AmoebaNet-A~\cite{Real2018Regularized}	 & 5.1M		   &    555M			& 74.5 / 92.0	 & EA\\
	AmoebaNet-B~\cite{Real2018Regularized}	 & 5.3M		   &    555M			& 74.0 / 91.5 	   & EA\\
	AmoebaNet-C~\cite{Real2018Regularized}	 & 5.1M		   &    535M			& 75.1 / 92.1	   & EA\\
	AmoebaNet-C (more filters)~\cite{Real2018Regularized}	 & 6.35M		   &    570M			& 75.7 / 92.4	 & EA\\
	PNASNet-5~\cite{liu2017progressive}		 & 5.1M		   &	 588M			& 74.2 / 91.9	  & SMBO\\
	ENAS~\cite{pham2018efficient}$^*$		  &	5.1M						&	  523M				& 74.3 / 91.9 	 & RL\\
	DARTS~\cite{DBLP:journals/corr/abs-1806-09055}		  &	4.9M		&	  595M					& 73.1 / 91.0 	 & Gradient\\

      \hline
	RENASNet (4, 44)					 & 5.36M		   &	 580M				& \textbf{75.7} / \textbf{92.6}	   & EA\&RL\\
      \hline
  \end{tabular}

    \begin{tablenotes}
 	\item[*] {\small{The result of ENAS was obtained by training with our setup, as it is not reported~\cite{pham2018efficient}.}}
  \end{tablenotes}
    \end{threeparttable}
    
  \end{table*}
  
\subsection{Implementation Details}
\subsubsection{Datasets Details}
\noindent
\textbf{CIFAR-10} \quad
CIFAR-10~\cite{Krizhevsky2009Learning} consists of 50,000 training images and 10,000 test images. 5,000 images are partitioned from the training set as a validation set. 
All images are whitened with the channel mean subtracted and the channel standard deviation divided. Then, we crop 32 x 32 patches from images and pad them to 40 x 40. These patches are also randomly fliped horizontally for data augmentation. When retraining the result architecture, we also use the cutout augmentation~\cite{cutout}.

\noindent
\textbf{ImageNet} \quad
For data augmentation on ImageNet~\cite{imagenet}, we resize the original input images with
its shorter side randomly sampled in [256, 480] for scale augmentation~\cite{simonyan2014very}.  $224 \times 224$ patches are randomly cropped from images.
Other standard operations, i.e., horizontal flip, mean pixel subtraction and the standard color augmentation in Alexnet~\cite{krizhevsky2012imagenet}, are also conducted~\cite{krizhevsky2012imagenet}. For the last 20 epochs, we withdraw most augmentations and only keep crop and flip augumentations for fine-tuning.

\subsubsection{Training details}
\noindent
\textbf{CIFAR-10} \quad
When training models on CIFAR-10, we use standard SGD optimizer with momentum rate set to 0.9, auxiliary classifier located at $\frac{2}{3}$ of the maximum depth weighted by 0.4, weight decay $3\times10^{-4}$, and dropout of 0.5 in the final softmax layer. In addition, we drop each path with probability 0.5 for regularization. Our batch size is 64 on each GPU and 2 GPUs are used. The learning rate initially is set to 0.05 and later decays with a cosine restart schedule for 630 epochs. 

\noindent
\textbf{ImageNet} \quad
When training models on ImageNet, we train each model for 200 epochs, using standard SGD optimizer with momentum rate set to 0.9, auxiliary classifier located at $\frac{2}{3}$ of the maximum depth weighted by 0.4, weight decay $4\times10^{-5}$. Our batch size is 64 on each GPU and 4 GPUs are used. The learning rate is initially set to 0.1 and later decays in a polynomial schedule.

\subsubsection{Details of the Controller}
For our controller, we use an LSTM with an embedding layer. The embedding size and the hidden state size of LSTM are both 100. The parameters of our controller are initialized with random values sampled from a normal distribution with a mean of zero and standard deviation of 0.01 and trained with Adam at a learning rate of 0.001. We apply a tanh constant of 2.5 and a temperature of 5.0 to the logits of the controller, and add the entropy of the controller to the reward with 0.1 weighted.

\subsubsection{Details of architecture search space}
For fair comparison, some details of our search space follow the NASNet search space: 
\begin{itemize}
\item[(1)] All convolutions follow an ordering of ReLU, convolution and batch normalization. 
\item[(2)] Each separable convolution is applied twice sequentially to the input feature map. 
\item[(3)] To match shapes in convolutional cells, 1x1 convolutions are applied as necessary. 
\item[(4)] Separable convolutions do not employ batch normalization or ReLU between depthwise and pointwise convolutions.
\end{itemize}

\subsection{Image Classification}
\label{5.2}

\subsubsection{Results on CIFAR-10}

Here we report the performance of our searched model, RENASNet, and make comparisons to other state-of-the-art models in Table~\ref{tab:1} on CIFAR-10. 
After the cell structures are fixed, we construct the entire networks same to the structure setting of NASNet\cite{zoph2017learning} and train them with the details mentioned before. The simple notation (6, 32) denotes cell unroll for $N=6$ times and $F=32$ filters in the first cell. 
 
\noindent
 The CIFAR-10 results are presented in Table~\ref{tab:1}. RENASNet achieves a competitive result to other state-of-the-art models. Only NASNet-A and AmoebaNet have stable better performances than RENASNet while they use much more computational resources (2000 GPU days and 3150 GPU days) than ours. ENAS is more efficient than our method, but our model has less parameters with better accuracy. 

\subsubsection{Results on ImageNet}
State-of-the-art image classifiers on ImageNet is shown in Table~\ref{tab:2}.
We conduct the comparison in the mobile setting where the image size is 224x224 and the multi-add operation numbers of models are under 600M. Note that as the accuracy of ENAS\cite{pham2018efficient} on ImageNet is not reported in the original paper, we trained it with all hyper-parameters and settings exactly same to RENASNet.

\noindent
The results on ImageNet are more convincing because CIFAR-10 is small and easy to be over-fitting.
The results on ImageNet are shown in Table~\ref{tab:2}. RENASNet surpasses both the manually designed models, including MobileNets~\cite{howard2017mobilenets,sandler2018inverted} and ShuffleNets~\cite{zhang2017shufflenet,DBLP:journals/corr/abs-1807-11164}, and the other state-of-the-art NAS models. Especially for NASNet~\cite{zoph2017learning} and AmoebaNet~\cite{Real2018Regularized}, they are representative RL-based and EA-based methods respectively and spend much more GPUs and days than ours.
In Table~\ref{tab:1} and Table~\ref{tab:2}, we also compare with DARTS~\cite{DBLP:journals/corr/abs-1806-09055}, which is a novel and gradient-based method. RENASNet is similar to DARTS~\cite{DBLP:journals/corr/abs-1806-09055} on CIFAR-10, but outperforms it on ImageNet.

\subsection{Search Efficiency}
\label{5.3}
Honestly speaking, although the computation cost of RENAS is much less than NASNet and AmoebaNet, it can not reflect a fair comparison of search efficiency to RL-based and EA-based NAS methods. As stated in Section~\ref{3}, the search space used in our experiments is smaller than the original NASNet search space. We have not distinguished Normal Cell and Reduction Cell and there are 6 operation choices. Thus, the efficiency of RENAS also comes from the search space.

\begin{figure}[t]
		\centering
		\includegraphics[width=0.5\textwidth]{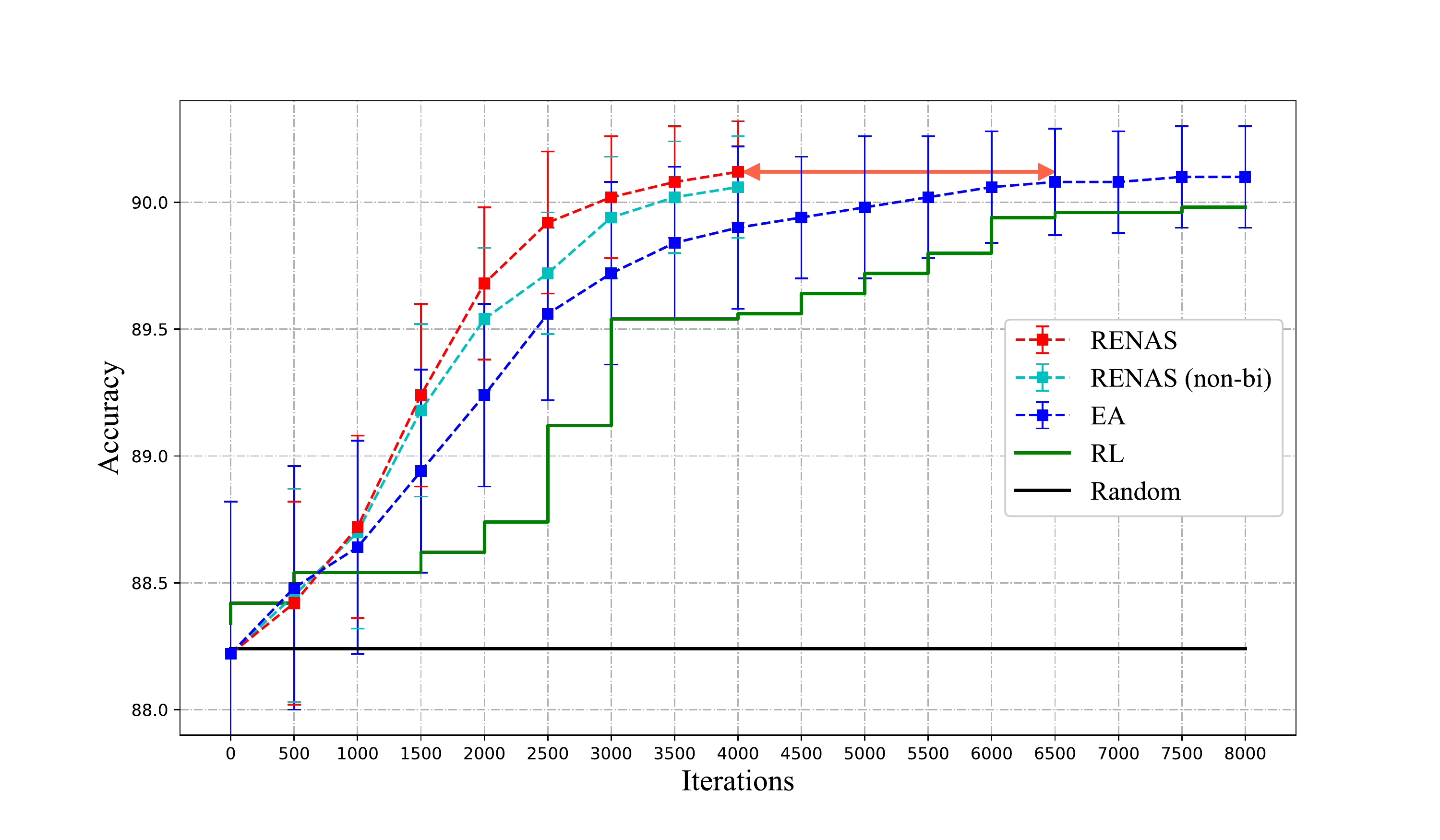}
		\captionof{figure}{Efficiency Comparison under the same search space (unified Cell and 6 operation choices).}
		\label{fig:4}
	\end{figure}

In this section, we make a fair ablation study to compare the efficiency of RENAS to EA and RL under the same search space (unified Cell and 6 operation choices) as in Fig.~\ref{fig:4}. For the compared methods, we keep track of the searched models for every 500 iterations. All searched models are evaluated with 20 epochs training from scratch. EA and RENAS are evaluated by the accuracy mean and variance of models in the population. EA is conduct with the same settings to RENAS, except that mutation actions are made randomly. RL is evaluated on the best model over time. Random is a model randomly picked from the search space. As shown in the Fig.~\ref{fig:4}, RENAS achieves better efficiency than EA and RL. The speedup of RENAS over RL/EA is around 1.5 - 2.0 times.

We also make an additional comparison. As mentioned in Section~\ref{4.2}, the controller is equipped with a bidirectional recurrent network to have a better capacity in architecture encoding. We compare RENAS to a counterpart with common recurrent network, RENAS (non-bi). It has exact same settings to RENAS, except the recurrent network. Fig.~\ref{fig:4} shows the inferiority of RENAS (non-bi).

		\begin{figure*}[ht]
		\centering
		\includegraphics[width=1.0\textwidth]{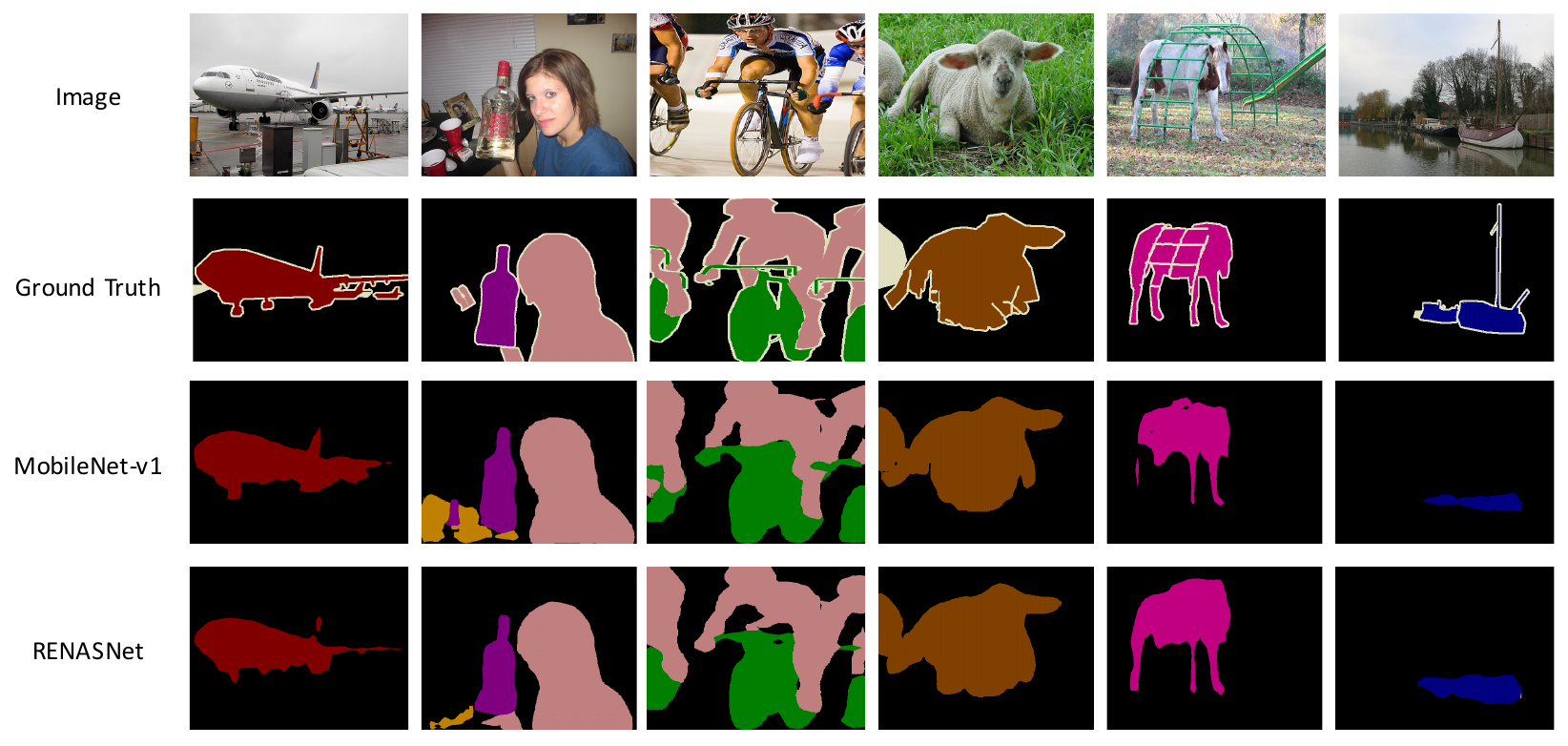}
		\captionof{figure}{Segmentation Results Visualization}
		\label{fig:5}
	\end{figure*}

\subsection{Semantic Segmentation}

  \begin{table}[!hbt]
  \centering
  \caption{Semantic Segmentation: DeepLabv3 on the PASCAL VOC 2012 validation set.} \label{tab:4}
      \begin{threeparttable}
  \begin{tabular}{ l  c | c  c  c }
  \hline
      Model 					& Pretrain & \#Params  &  mIOU(\%) \\ 
      \hline  
	MobileNet-v1 \cite{howard2017mobilenets}	 & COCO		 & 11.15M	  & 75.29  \\ 
	MobileNet-v2 \cite{sandler2018inverted}		 	 & COCO		 &  4.52M		   & 75.70	\\ 
      \hline
	MobileNet-v1 \cite{howard2017mobilenets}	 & ImageNet		 & 11.14M   & 68.79   \\ 
	MobileNet-v2 \cite{sandler2018inverted}		 	 & ImageNet		 & 4.51M	   & 70.02	\\  
	NASNet-A \cite{zoph2017learning}		 			 & ImageNet		 &	 12.39M   & 73.68	\\ 
	RENASNet 					 										 & ImageNet		 &	 11.63M   & \textbf{75.83}	\\
      \hline
  \end{tabular}
    \end{threeparttable}
  \end{table}
  
In this section, we make further experiments on semantic segmentation with DeepLabv3~\cite{deeplab-v3}. All our experiments and comparison methods use Atrous Spatial Pyramid Pooling module (ASPP)~\cite{deeplab-v2} that contains three 3x3 convolutions with different atrous rates. 
The output stride is 16 that is the ratio of input image spatial resolution to final output resolution. We do not use Multi-scale and left-right Flipped inputs (MF), which is employed by some other works~\cite{sandler2018inverted} for boosting the performance. Following our comparison methods, we conduct the experiments on the PASCAL VOC 2012 dataset~\cite{PascalVOC} and standard extra annotated images from \cite{extraImage} with evaluation metric as mIOU.
 
 We compare RENASNet with three other mobile networks, MobileNet-v1~\cite{howard2017mobilenets}, MobileNet-v2~\cite{sandler2018inverted} and NASNet-A~\cite{zoph2017learning} and summarize the results in Table~\ref{tab:4}. The results of models pretrained on COCO~\cite{COCO} are reported in~\cite{sandler2018inverted}. Models pretrained on ImageNet are implemented by ourselves using exactly same hyper-parameters and settings.
 From the results, we have observed that: 1) The performance of this task relies heavily on pretrained models.  Without being pretrained on COCO, MobileNet-v1 and MobileNet-v2 suffer from severe performance decay. 2) In terms of mIOU, RENASNet outperforms MobileNet-v1, MobileNet-v2 and NASNet-A~\cite{howard2017mobilenets} at the same settings. Moreover, RENASNet~(75.83\%) pretrained on ImageNet even performs better than MobileNet-v1~(75.29\%) and MobileNet-v2~(75.70\%) that pretrained on COCO. The segmentation results are visualized in Fig.~\ref{fig:5}.

\section{Conclusion}
In this paper, we have proposed a method for neural architecture search by integrating evolution algorithm and reinforcement learning into a unified framework. Inspired by the procedure of designing networks manually, we use a controller to learn the effects of modifications and make better mutation actions. The searched architecture, i.e., RENASNet, achieves competitive performance on CIFAR-10 and outperforms other state-of-the-art models on ImageNet (75.7\% top-1 accuracy with 5.36M parameters). 
In addition, RENASNet also demonstrate its high performance on the semantic segmentation task. RENASNet outperforms other mobile size networks and achieves 75.83\% mIOU without being pretrained on COCO. It shows that RENASNet can be transferred to other computer vision tasks in addition to image classification. In future, we will try to conduct NAS on other tasks, e.g., object detection.

\section*{Acknowledgement}
 This work was supported by the National Natural Science Foundation of China under Grants 91646207, 61773377,  61573352 and 61876212, and the Beijing Natural Science Foundation under Grant L172053.

{\small
\bibliographystyle{ieee}
\bibliography{egbib}

\begin{thebibliography}{10}\itemsep=-1pt

\bibitem{baker2016designing}
B.~Baker, O.~Gupta, N.~Naik, and R.~Raskar.
\newblock Designing neural network architectures using reinforcement learning.
\newblock {\em CoRR}, abs/1611.02167, 2016.

\bibitem{baker2018accelerating}
B.~Baker, O.~Gupta, R.~Raskar, and N.~Naik.
\newblock Accelerating neural architecture search using performance prediction.
\newblock {\em CoRR}, abs/1705.10823, 2017.

\bibitem{cai2018efficient}
H.~Cai, T.~Chen, W.~Zhang, Y.~Yu, and J.~Wang.
\newblock Efficient architecture search by network transformation.
\newblock In {\em AAAI}, pages 2787--2794, 2018.

\bibitem{deeplab-v2}
L.~Chen, G.~Papandreou, I.~Kokkinos, K.~Murphy, and A.~L. Yuille.
\newblock Deeplab: Semantic image segmentation with deep convolutional nets,
  atrous convolution, and fully connected crfs.
\newblock {\em {IEEE} Trans. Pattern Anal. Mach. Intell.}, 40(4):834--848,
  2018.

\bibitem{deeplab-v3}
L.~Chen, G.~Papandreou, F.~Schroff, and H.~Adam.
\newblock Rethinking atrous convolution for semantic image segmentation.
\newblock {\em CoRR}, abs/1706.05587, 2017.

\bibitem{chen2015net2net}
T.~Chen, I.~J. Goodfellow, and J.~Shlens.
\newblock Net2net: Accelerating learning via knowledge transfer.
\newblock {\em CoRR}, abs/1511.05641, 2015.

\bibitem{imagenet}
J.~Deng, W.~Dong, R.~Socher, L.~Li, K.~Li, and F.~Li.
\newblock Imagenet: {A} large-scale hierarchical image database.
\newblock In {\em CVPR}, pages 248--255, 2009.

\bibitem{cutout}
T.~Devries and G.~W. Taylor.
\newblock Improved regularization of convolutional neural networks with cutout.
\newblock {\em CoRR}, abs/1708.04552, 2017.

\bibitem{downing2001adaptive}
K.~L. Downing.
\newblock Adaptive genetic programs via reinforcement learning.
\newblock In {\em GEC}, 2001.

\bibitem{DBLP:journals/gpem/Downing01}
K.~L. Downing.
\newblock Reinforced genetic programming.
\newblock {\em Genetic Programming and Evolvable Machines}, 2(3):259--288,
  2001.

\bibitem{PascalVOC}
M.~Everingham, S.~M.~A. Eslami, L.~J.~V. Gool, C.~K.~I. Williams, J.~M. Winn,
  and A.~Zisserman.
\newblock The pascal visual object classes challenge: {A} retrospective.
\newblock {\em International Journal of Computer Vision}, 111(1):98--136, 2015.

\bibitem{goldberg1991comparative}
D.~E. Goldberg and K.~Deb.
\newblock A comparative analysis of selection schemes used in genetic
  algorithms.
\newblock In {\em FGA}, pages 69--93. 1990.

\bibitem{extraImage}
B.~Hariharan, P.~Arbelaez, L.~D. Bourdev, S.~Maji, and J.~Malik.
\newblock Semantic contours from inverse detectors.
\newblock pages 991--998, 2011.

\bibitem{he2016deep}
K.~He, X.~Zhang, S.~Ren, and J.~Sun.
\newblock Deep residual learning for image recognition.
\newblock In {\em CVPR}, pages 770--778, 2016.

\bibitem{hitoshi1998multi}
I.~Hitoshi.
\newblock Multi-agent reinforcement learning with genetic programming.
\newblock In {\em GP}, 1998.

\bibitem{howard2017mobilenets}
A.~G. Howard, M.~Zhu, B.~Chen, D.~Kalenichenko, W.~Wang, T.~Weyand,
  M.~Andreetto, and H.~Adam.
\newblock Mobilenets: Efficient convolutional neural networks for mobile vision
  applications.
\newblock {\em CoRR}, abs/1704.04861, 2017.

\bibitem{huang2017densely}
G.~Huang, Z.~Liu, L.~van~der Maaten, and K.~Q. Weinberger.
\newblock Densely connected convolutional networks.
\newblock In {\em CVPR}, pages 2261--2269, 2017.

\bibitem{kamio2005adaptation}
S.~Kamio and H.~Iba.
\newblock Adaptation technique for integrating genetic programming and
  reinforcement learning for real robots.
\newblock {\em {IEEE} Trans. Evolutionary Computation}, 9(3):318--333, 2005.

\bibitem{Krizhevsky2009Learning}
A.~Krizhevsky and G.~Hinton.
\newblock Learning multiple layers of features from tiny images.
\newblock {\em Technical report.}, 1(4):1--7, 2009.

\bibitem{krizhevsky2012imagenet}
A.~Krizhevsky, I.~Sutskever, and G.~E. Hinton.
\newblock Imagenet classification with deep convolutional neural networks.
\newblock In {\em NIPS}, pages 1106--1114, 2012.

\bibitem{COCO}
T.~Lin, M.~Maire, S.~J. Belongie, J.~Hays, P.~Perona, D.~Ramanan,
  P.~Doll{\'{a}}r, and C.~L. Zitnick.
\newblock Microsoft {COCO:} common objects in context.
\newblock In {\em {ECCV}}, pages 740--755, 2014.

\bibitem{liu2017progressive}
C.~Liu, B.~Zoph, M.~Neumann, J.~Shlens, W.~Hua, L.~Li, L.~Fei{-}Fei, A.~L.
  Yuille, J.~Huang, and K.~Murphy.
\newblock Progressive neural architecture search.
\newblock In {\em ECCV}, pages 19--35, 2018.

\bibitem{DBLP:journals/corr/abs-1806-09055}
H.~Liu, K.~Simonyan, and Y.~Yang.
\newblock {DARTS:} differentiable architecture search.
\newblock {\em CoRR}, abs/1806.09055, 2018.

\bibitem{DBLP:journals/corr/abs-1807-11164}
N.~Ma, X.~Zhang, H.~Zheng, and J.~Sun.
\newblock Shufflenet {V2:} practical guidelines for efficient {CNN}
  architecture design.
\newblock In {\em ECCV}, pages 122--138, 2018.

\bibitem{miikkulainen2017evolving}
R.~Miikkulainen, J.~Z. Liang, E.~Meyerson, A.~Rawal, D.~Fink, O.~Francon,
  B.~Raju, H.~Shahrzad, A.~Navruzyan, N.~Duffy, and B.~Hodjat.
\newblock Evolving deep neural networks.
\newblock {\em CoRR}, abs/1703.00548, 2017.

\bibitem{miller1989designing}
G.~F. Miller, P.~M. Todd, and S.~U. Hegde.
\newblock Designing neural networks using genetic algorithms.
\newblock In {\em ICGA}, pages 379--384, 1989.

\bibitem{pham2018efficient}
H.~Pham, M.~Y. Guan, B.~Zoph, Q.~V. Le, and J.~Dean.
\newblock Efficient neural architecture search via parameter sharing.
\newblock In {\em ICML}, pages 4092--4101, 2018.

\bibitem{Real2018Regularized}
E.~Real, A.~Aggarwal, Y.~Huang, and Q.~V. Le.
\newblock Regularized evolution for image classifier architecture search.
\newblock {\em CoRR}, abs/1802.01548, 2018.

\bibitem{real2017large}
E.~Real, S.~Moore, A.~Selle, S.~Saxena, Y.~L. Suematsu, J.~Tan, Q.~V. Le, and
  A.~Kurakin.
\newblock Large-scale evolution of image classifiers.
\newblock In {\em ICML}, pages 2902--2911, 2017.

\bibitem{sandler2018inverted}
M.~Sandler, A.~G. Howard, M.~Zhu, A.~Zhmoginov, and L.~Chen.
\newblock Inverted residuals and linear bottlenecks: Mobile networks for
  classification, detection and segmentation.
\newblock {\em CoRR}, abs/1801.04381, 2018.

\bibitem{simonyan2014very}
K.~Simonyan and A.~Zisserman.
\newblock Very deep convolutional networks for large-scale image recognition.
\newblock {\em CoRR}, abs/1409.1556, 2014.

\bibitem{stanley2002evolving}
K.~O. Stanley and R.~Miikkulainen.
\newblock Evolving neural networks through augmenting topologies.
\newblock {\em Evolutionary computation}, 10(2):99--127, 2002.

\bibitem{szegedy2017inception}
C.~Szegedy, S.~Ioffe, V.~Vanhoucke, and A.~A. Alemi.
\newblock Inception-v4, inception-resnet and the impact of residual connections
  on learning.
\newblock In {\em AAAI}, pages 4278--4284, 2017.

\bibitem{szegedy2015going}
C.~Szegedy, W.~Liu, Y.~Jia, P.~Sermanet, S.~E. Reed, D.~Anguelov, D.~Erhan,
  V.~Vanhoucke, and A.~Rabinovich.
\newblock Going deeper with convolutions.
\newblock In {\em CVPR}, pages 1--9, 2015.

\bibitem{szegedy2016rethinking}
C.~Szegedy, V.~Vanhoucke, S.~Ioffe, J.~Shlens, and Z.~Wojna.
\newblock Rethinking the inception architecture for computer vision.
\newblock In {\em CVPR}, pages 2818--2826, 2016.

\bibitem{xie2017genetic}
L.~Xie and A.~L. Yuille.
\newblock Genetic {CNN}.
\newblock In {\em ICCV}, pages 1388--1397, 2017.

\bibitem{DBLP:journals/tnn/YaoL97}
X.~Yao and Y.~Liu.
\newblock A new evolutionary system for evolving artificial neural networks.
\newblock {\em {IEEE} Trans. Neural Networks}, 8(3):694--713, 1997.

\bibitem{zhang2017shufflenet}
X.~Zhang, X.~Zhou, M.~Lin, and J.~Sun.
\newblock Shufflenet: An extremely efficient convolutional neural network for
  mobile devices.
\newblock {\em CoRR}, abs/1707.01083, 2017.

\bibitem{zhong2018practical}
Z.~Zhong, J.~Yan, W.~Wu, J.~Shao, and C.-L. Liu.
\newblock Practical block-wise neural network architecture generation.
\newblock In {\em CVPR}, 2018.

\bibitem{zoph2016neural}
B.~Zoph and Q.~V. Le.
\newblock Neural architecture search with reinforcement learning.
\newblock {\em CoRR}, abs/1611.01578, 2016.

\bibitem{zoph2017learning}
B.~Zoph, V.~Vasudevan, J.~Shlens, and Q.~V. Le.
\newblock Learning transferable architectures for scalable image recognition.
\newblock {\em CoRR}, abs/1707.07012, 2017.

\end{thebibliography}
}

\end{document}